\documentclass{article}
\usepackage[final,dblblindworkshop]{neurips_2025}
\workshoptitle{Mechanistic Interpretability}
\usepackage{paralist}
\usepackage[utf8]{inputenc}
\usepackage[T1]{fontenc}
\usepackage{amsmath, amssymb, amsthm}
\usepackage{graphicx}
\usepackage[colorlinks=true,urlcolor=blue,linkcolor=.,citecolor=.,filecolor=.,menucolor=.]{hyperref}
\usepackage{booktabs}
\usepackage{enumitem}
\usepackage{listings}
\usepackage{xcolor}
\usepackage{xspace}
\usepackage{cleveref}

\def\nnterp{\texttt{nnterp}\xspace}
\title{\nnterp: A Standardized Interface for Mechanistic Interpretability of Transformers}
\author{
  Cl\'ement Dumas\\
  MATS, \'Ecole Normale Sup\'erieure Paris-Saclay, Universit\'e Paris-Saclay\\
  \texttt{clement.dumas@ens-paris-saclay.fr}
  }

\date{}

\begin{document}
\maketitle

\begin{abstract}
Mechanistic interpretability research requires reliable tools for analyzing transformer internals across diverse architectures. Current approaches face a fundamental tradeoff: custom implementations like TransformerLens ensure consistent interfaces but require coding a manual adaptation for each architecture, introducing numerical mismatch with the original models, while direct HuggingFace access through NNsight preserves exact behavior but lacks standardization across models. To bridge this gap, we develop \nnterp, a lightweight wrapper around NNsight that provides a unified interface for transformer analysis while preserving original HuggingFace implementations. Through automatic module renaming and comprehensive validation testing, \nnterp enables researchers to write intervention code once and deploy it across 50+ model variants spanning 16 architecture families. The library includes built-in implementations of common interpretability methods (logit lens, patchscope, activation steering) and provides direct access to attention probabilities for models that support it. By packaging validation tests with the library, researchers can verify compatibility with custom models locally. \nnterp bridges the gap between correctness and usability in mechanistic interpretability tooling.\footnote{Repository available at \url{https://github.com/Butanium/nnterp}, documentation at \url{https://butanium.github.io/nnterp/}. Install with \texttt{pip install nnterp}.}
\end{abstract}

\begin{figure}[t]
    \centering
    \includegraphics[width=\textwidth]{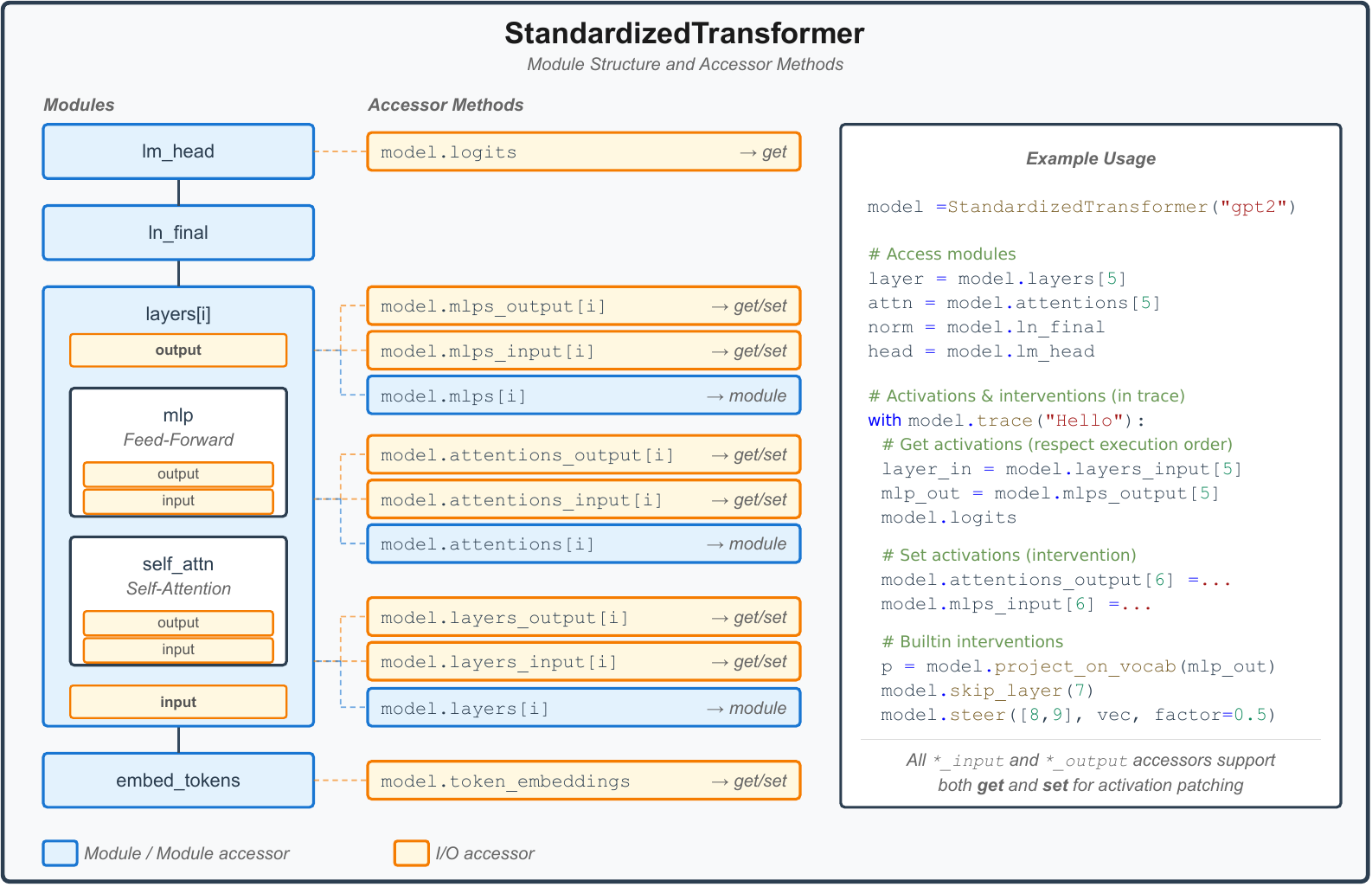}
    \caption{
        \nnterp provides a standardized interface for transformer models.
        \textbf{Left:} It renames transformer modules to a consistent naming scheme (\texttt{layers}, \texttt{self\_attn}, \texttt{mlp}, etc.).
        \textbf{Middle:} It provides accessor methods for internal activations, with \texttt{get} and \texttt{set} for all \texttt{*\_input} and \texttt{*\_output}. This handles whether the module returns a tuple or a single tensor.
        \textbf{Right:} Example usage for intervention and analysis.
    }
    \label{fig:nnterp-structure}
\end{figure}

\section{Introduction}

Mechanistic interpretability research aims to reverse-engineer the computational mechanisms within neural networks \citep{elhage2021mathematical, olah2020zoom}. For transformer language models, this requires tools that can reliably access and modify internal representations across diverse architectures. However, the field faces a fundamental engineering challenge: balancing implementation correctness with practical usability across the rapidly expanding landscape of transformer variants.

Two dominant paradigms have emerged. TransformerLens \citep{nanda2022transformerlens} provides a clean, unified interface by reimplementing transformer architectures from scratch. This ensures consistent module naming and reliable hooks but requires manual implementation for each new architecture, may introduce subtle differences from original models, and cannot leverage architecture-specific optimizations. Conversely, NNsight \citep{fiottokaufman2024nnsightndifdemocratizingaccess} operates directly on HuggingFace implementations, preserving exact model behavior and supporting any architecture HuggingFace provides. However, this approach inherits the fragmentation of HuggingFace's naming conventions—accessing layers requires \texttt{model.transformer.h} for GPT-2 but \texttt{model.model.layers} for LLaMA—and leaves researchers vulnerable to breaking changes in transformer implementations. For example, since HugginFace transformers 4.54, Qwen and Llama layers return activation tensors instead of tuples, which caused silent bugs in many interpretability experiments.

This fragmentation creates significant friction. Researchers must either commit to a single tool's limitations or maintain parallel codebases for different architectures. As mechanistic interpretability increasingly emphasizes cross-model validation and scalability to large models requiring optimizations, neither approach alone suffices.

We present \nnterp, a lightweight library that bridges this gap by providing a standardized interface atop NNsight's HuggingFace integration. Through systematic module renaming and validation, \nnterp enables researchers to write \texttt{model.layers\_output[5]} consistently across GPT-2, LLaMA, Gemma, and other architectures while preserving exact HuggingFace behavior. Key contributions include:

\begin{compactitem}
\item A unified API for accessing transformer internals (layers, attention, MLP outputs) that works identically across 50+ model variants from 16 architecture families.
\item Automatic validation upon model loading that verifies module shapes and intervention correctness. This caught the HuggingFace transformers 4.54 bug mentioned earlier day-1.
\item Built-in implementations of common interpretability methods (logit lens, patchscope, steering) that work across all supported models.
\item Direct access to attention probabilities for compatible architectures through NNsight's intermediate variable tracking.
\item Packaged test suite enabling local verification of custom models.
\end{compactitem}

\section{Background and Related Work}

\paragraph{Mechanistic Interpretability Tooling.}
Early mechanistic interpretability relied on manual PyTorch hooks to intercept activations \citep{elhage2021mathematical}. As the field matured, specialized libraries emerged to streamline common operations. Pyvene \citep{wu2024pyvene} provides a declarative framework for causal interventions, while TransformerLens \citep{nanda2022transformerlens} offers a unified interface through custom implementations. NNsight \citep{fiottokaufman2024nnsightndifdemocratizingaccess} takes a different approach, wrapping existing models with a tracing system that enables interventions while preserving original implementations.

\paragraph{The Standardization Challenge.}
HuggingFace Transformers \citep{wolf2020transformers} has become the de facto repository for transformer implementations, but its design prioritizes flexibility over consistency. Each architecture follows its own naming conventions and internal structure. This heterogeneity reflects genuine architectural differences but complicates systematic analysis. Recent work on model diffing \citep{lindsey2024sparse} and circuit discovery \citep{conmy2023automatedcircuit} highlights the need for tools that work reliably across architectures.

\section{nnterp Design and Implementation}

\nnterp extends NNsight's \texttt{LanguageModel} class with a \texttt{StandardizedTransformer} that automatically renames modules to follow a consistent naming scheme and provides some I/O accessors as shown in \Cref{fig:nnterp-structure}. The module renaming system is implemented using the \texttt{rename} argument from the \texttt{NNsight} class, which allow to pass a dictionary of original names and new names. Different architectures require different renaming strategies. GPT-2's \texttt{transformer.h}, \texttt{attn}, \texttt{transformer.ln\_f} becomes \texttt{layers}, \texttt{self\_attn}, \texttt{ln\_final}, while LLaMA's structure just moves \texttt{model.layers} to \texttt{layers}. \nnterp maintains a configuration system that maps each architecture class to its renaming rules:
\begin{lstlisting}[language=Python, basicstyle=\small,  breaklines=true]
model = StandardizedTransformer("gpt2")
# Internally applies:
rename = {
  "transformer": "model", "h": "layers", "model.layers": "layers",
  "attn": "self_attn", "transformer.ln_f": "ln_final"
}
\end{lstlisting}

\nnterp also provides \texttt{model.\{layers/mlps/attentions\}\_{input/output}[layer\_idx]} which allow to get and set the input and output of the specified layer. This should be preferred over using e.g. \texttt{model.layers[layer\_idx].output} as this can be a tensor or a tuple depending on the architecture, while \texttt{model.layers\_output[layer\_idx]} always get/set the output tensor.

This standardization leverages NNsight's module renaming capability while maintaining full compatibility with NNsight's intervention API (the modules are still available under their original names). Researchers can use \nnterp's simplified interface for common operations or drop down to raw NNsight for advanced use cases.

\paragraph{Attention probabilities access.}
To enable access to attention probabilities -- which requires to use the slower eager attention implementation -- \texttt{enable\_attention\_probs=True} needs to be passed to the \texttt{StandardizedTransformer} class. Once enabled, the attention probabilities can be accessed and set using the \texttt{model.attention\_probabilities[layer\_idx]} property.
This relies on NNsight's \texttt{source} feature, which allows to access intermediate variables in the forward pass. This means, that this is very implementation sensitive, and may break with new HuggingFace releases, e.g. if the name of the attention variable changes. The attention probabilities hookpoint is therefore not available on all models. See \Cref{app:add_support} for how to add support for new models.

\paragraph{Validation guarantees.}
Upon initialization, \nnterp runs automatic tests to verify: (1) module outputs have expected shapes, (2) attention probabilities sum to 1, (3) interventions affect outputs, and (4) layer skip operations preserve causality. These tests catch common issues like incorrect module identification or incompatible attention implementations. The validation suite ships with the package, allowing researchers to test custom models locally via \texttt{python -m nnterp run\_tests}.

\section{Extra features}
\paragraph{Built-in interventions.}
\nnterp implements common interpretability methods that work across all supported models. \textbf{Logit Lens} \citep{nostalgebraist2020logitlens} projects hidden states through the unembedding to see intermediate predictions. \textbf{Patchscope} \citep{ghandeharioun2024patchscope} replaces activations from one context into another. \textbf{Activation Steering} adds steering vectors at specified layers. All methods use the same unified API across architectures.

\paragraph{Prompt Management.}
\nnterp provides a \texttt{Prompt} class that allows to track probabilities of specific target tokens in different categories. The easiest way to create a \texttt{Prompt} is to use the \texttt{Prompt.from\_strings} method, which takes a prompt string and a dictionary of categories and target strings. The tracked tokens of each category is the set of all first tokens of the target strings with and without beginning of word. E.g for \texttt{["London", "Lyon"]}, the tracked tokens could be \texttt{["\_London", "Lon", "\_Lyo", "Ly"]}.
\begin{lstlisting}[language=Python, basicstyle=\small]
prompt = Prompt.from_strings(
    "The capital of France is",
    {"target": "Paris", "fake": ["London", "Lyon"]},
    tokenizer
)
results = run_prompts(model, [prompt])
# Returns probabilities for each category
{"target": torch.Tensor([0.7]), "fake": torch.Tensor([0.1])}
\end{lstlisting}

\section{Empirical Validation}

\nnterp supports 21 architecture families\footnote{See \Cref{app:model_coverage} for the full list.}.
\nnterp adds minimal overhead to NNsight's already efficient implementation. NNsight's performance analysis \citep{fiottokaufman2024nnsightndifdemocratizingaccess} shows it matches or exceeds TransformerLens speed while using less memory. Since \nnterp is a thin wrapper providing only interface standardization, it inherits these performance characteristics.

\section{Discussion and Future Work}

\paragraph{Limitations.}
\nnterp's validation tests provide sanity checks rather than formal correctness guarantees. While they catch common issues, subtle bugs in attention probability hooks or module identification may persist. The library also inherits NNsight's limitations, including incompatibility with some attention implementations (e.g., Flash Attention for attention probabilities).

\paragraph{Impact on Research Workflow.}
By separating interface standardization from implementation details, \nnterp enables more reproducible mechanistic interpretability research. Researchers can share intervention code that works across models, facilitating replication and cross-architecture validation. The packaged test suite ensures that even custom models can be verified for compatibility.

\paragraph{Future Directions.}
Future work includes automated architecture detection, support for non-causal and encoder-decoder architectures, access to attention KQV and MLP intermediate activations and access to MoE router's logits. We are also exploring integration with NNsight itself and how to support remote execution via NDIF, which allow researcher to run their NNsight experiments on remote machines.

\section{Conclusion}

\nnterp demonstrates that the tradeoff between implementation correctness and interface usability in mechanistic interpretability tooling is not fundamental. By leveraging NNsight's model wrapping capabilities and adding systematic validation, we provide researchers with both reliable access to exact HuggingFace implementations and a consistent interface across architectures. As mechanistic interpretability scales to larger models and broader architectural diversity, tools that balance correctness with usability become essential. \nnterp represents a step toward more robust and reproducible interpretability research.

\bibliographystyle{unsrtnat}
\bibliography{references}

\appendix
\crefalias{section}{appendix}

\section{Adding Support for Custom Models}\label{app:add_support}

\nnterp uses a standardized naming convention to provide a unified interface across transformer architectures. When a model doesn't follow the expected naming patterns, researchers can use \texttt{RenameConfig} to map custom module names to the standardized interface.

\subsection{Target Structure}

\nnterp expects all models to follow this structure:
\begin{lstlisting}[language=Python, basicstyle=\small]
StandardizedTransformer
|-- embed_tokens
|-- layers[i]
|   |-- self_attn
|   `-- mlp
|-- ln_final
`-- lm_head
\end{lstlisting}

Models are automatically renamed to match this pattern using built-in mappings. The library provides convenient accessors: \texttt{layers\_input[i]}, \texttt{layers\_output[i]} for layer I/O, \texttt{attentions[i]}, \texttt{attentions\_input[i]}, \texttt{attentions\_output[i]} for attention, and \texttt{mlps[i]}, \texttt{mlps\_input[i]}, \texttt{mlps\_output[i]} for MLP components.

\subsection{Basic RenameConfig Usage}

When automatic renaming fails, create a custom \texttt{RenameConfig} specifying module names in the original model:

\begin{lstlisting}[language=Python, basicstyle=\small]
from nnterp import StandardizedTransformer
from nnterp.rename_utils import RenameConfig

rename_config = RenameConfig(
    model_name="custom_transformer",
    layers_name="custom_layers",
    attn_name="custom_attention",
    mlp_name="custom_ffn",
    ln_final_name="custom_norm",
    lm_head_name="custom_head"
)

model = StandardizedTransformer(
    "your-model-name",
    rename_config=rename_config
)
\end{lstlisting}

For nested modules, use dot notation (e.g., \texttt{layers\_name="transformer.encoder\_layers"}). Multiple alternative names can be provided as lists (e.g., \texttt{attn\_name=["attention", "self\_attention"]}).

\subsection{Implementing Attention Probabilities}

Attention probabilities support requires implementing \texttt{AttnProbFunction}. Here's the GPT-J implementation:

\begin{lstlisting}[language=Python, basicstyle=\small]
from nnterp.rename_utils import AttnProbFunction

class GPTJAttnProbFunction(AttnProbFunction):
    def get_attention_prob_source(self, attention_module,
                                   return_module_source=False):
        if return_module_source:
            return attention_module.source.self__attn_0.source
        return attention_module.source.self__attn_0.source\
               .self_attn_dropout_0

model = StandardizedTransformer(
    "yujiepan/gptj-tiny-random",
    enable_attention_probs=True,
    rename_config=RenameConfig(
        attn_prob_source=GPTJAttnProbFunction()
    )
)
\end{lstlisting}

The process involves: (1) using \texttt{model.scan()} to explore the forward pass, (2) locating where attention probabilities are computed (typically after dropout), (3) implementing the hook via \texttt{AttnProbFunction}, and (4) testing with dummy inputs to verify shapes and normalization.

\subsection{Validation}

\nnterp automatically validates configurations during initialization, checking: module naming correctness, tensor shapes at each layer, attention probabilities normalization (if enabled), and I/O compatibility. Manual validation can be performed using \texttt{model.trace()} to verify expected tensor shapes and properties.

\section{Model coverage}\label{app:model_coverage}
The following model classes were tested and work with \nnterp:
\begin{itemize}
    \item \texttt{BloomForCausalLM}
    \item \texttt{BloomModel}
    \item \texttt{Ernie4\_5\_MoeForCausalLM}
    \item \texttt{GPT2LMHeadModel}
    \item \texttt{GPTBigCodeForCausalLM}
    \item \texttt{GPTJForCausalLM}
    \item \texttt{Gemma2ForCausalLM}
    \item \texttt{Gemma3ForCausalLM}
    \item \texttt{Gemma3ForConditionalGeneration}
    \item \texttt{GemmaForCausalLM}
    \item \texttt{Glm4ForCausalLM}
    \item \texttt{Glm4MoeForCausalLM}
    \item \texttt{LlamaForCausalLM}
    \item \texttt{MistralForCausalLM}
    \item \texttt{MixtralForCausalLM}
    \item \texttt{OPTForCausalLM}
    \item \texttt{Phi3ForCausalLM}
    \item \texttt{Qwen2ForCausalLM}
    \item \texttt{Qwen3ForCausalLM}
    \item \texttt{Qwen3MoeForCausalLM}
    \item \texttt{SeedOssForCausalLM}
    \item \texttt{SmolLM3ForCausalLM}
    \item \texttt{DbrxForCausalLM}
    \item \texttt{GptOssForCausalLM}
    \item \texttt{Qwen2MoeForCausalLM}
    \item \texttt{StableLmForCausalLM}
\end{itemize}

Support for attention probabilities is still a work in progress for the following model classes:
\begin{itemize}
    \item \texttt{DbrxForCausalLM}
    \item \texttt{GptOssForCausalLM}
    \item \texttt{Qwen2MoeForCausalLM}
    \item \texttt{StableLmForCausalLM}
\end{itemize}

\end{document}